**On the Challenges of Sentiment Analysis for Dynamic Events**

Monireh Ebrahimi[1], Amir Hossein Yazdavar[1], Amit Sheth[1,2]
[1] Kno.e.sis Center, Wright State University
[2] Cognovi Labs
{monireh, amir, amit}@knoesis.org

With the proliferation of social media over the last decade, determining people's attitude with respect to a specific topic, document, interaction or events has fueled research interest in natural language processing and introduced a new channel called "sentiment and emotion analysis" [1]. For instance, businesses routinely look to develop systems to automatically understand their customer conversations by identifying the relevant content to enhance marketing their products and managing their reputations [2]. Previous efforts to assess people's sentiment on Twitter have suggested that Twitter may be a valuable resource for studying political sentiment and that it reflects the offline political landscape. According to a Pew Research Center report, in January 2016 44% of US adults stated having learned about the presidential election through social media. Furthermore, 24% reported use of social media posts of the two candidates as a source of news and information, which is more than the 15% who have used both candidates' websites or emails combined (http://j.mp/PewSocM). The first presidential debate between Trump and Hillary was the most tweeted debate ever with 17.1 million tweets.

Many opinion mining systems and tools have been developed to provide users with the attitudes of people towards products/people/topics and their attributes/aspects. However, sentiment analysis for predicting the result of an election is still a challenging task. Though apparently simple, it is empirically highly challenging to train a successful model for conducting sentiment analysis on tweet streams for a dynamic event such as an election. Among the key challenges are changes in the topics of conversation and the people about which social media posts express their opinions. In this article, we will highlight some of the challenges related to sentiment analysis we encountered during our monitoring of the presidential election using Kno.e.sis' Twitris system [17]. Twitris has helped in several successful election predictions, including 2012 US Election [5], Brexit (http://tcrn.ch/29DNjon, http://bit.ly/TwBrexit), and the 2016 US Election (http://j.mp/E16Cognovi, http://bit.ly/Ele2016). The latter two involved collaboration between the Kno.e.sis Center and Cognovi Labs, a startup based on the Twitris technology for the primary purpose of evaluating how the technology scales for real-time analysis. Figure 1 shows a dashboard used for real-time monitoring and analysis.

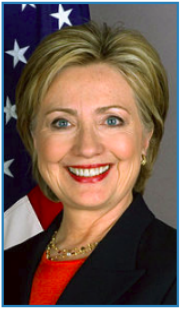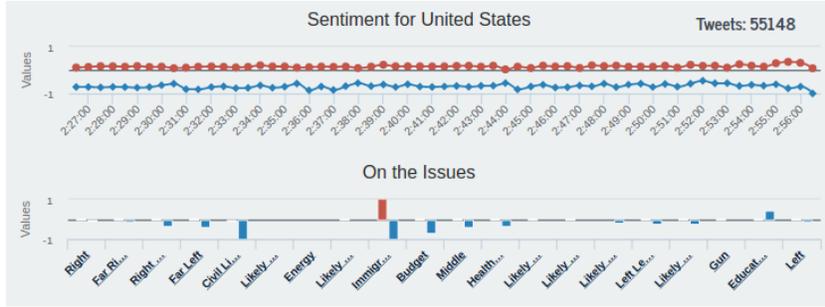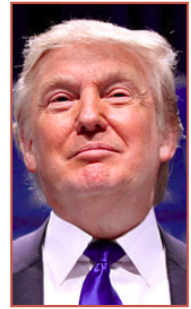
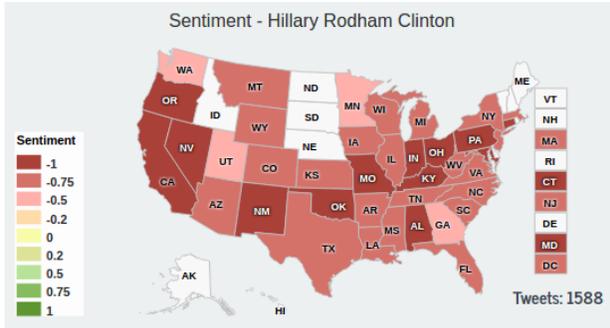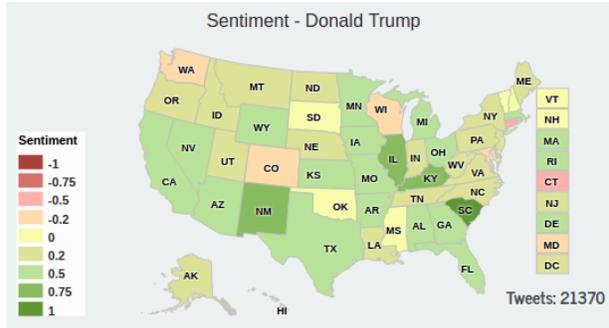

Figure 1. The Twitris system was used in real-time to analyze the three presidential debates and for prediction on election-day for 2016 US election.

We first created a supervised multi-class classifier (positive vs. negative vs. neutral) for analyzing opinions about different election candidates as expressed in the tweets. To this end, we trained our model for each candidate separately. The motivation for this segregation comes from our observation that the same tweet on an issue can be positive for one candidate while negative for another. In fact, the sentiment of a tweet is very candidate-dependent. In the first round of training in July 2016 before the convention, we used 10,000 labeled tweets collected for 5 candidates, including Bernie Sanders, Donald Trump, Hillary Clinton, John Kasich, and Ted Cruz, on 9 issues encompassing budget, finance, education, energy, environment, healthcare, immigration, gun control, and civil liberties. In addition to excluding re-tweets, tweets were tested for similarity using a ratio of Levenshtein distance to ensure that no two tweets were too similar. Afterward, through many experiments over different machine learning algorithms and parameter settings, we found our best model with respect to F-measure. Our best traditional machine learning-based model for Clinton uses SVM with TF-IDF vectorization of 1-3 grams, positive and negative hashtags for each candidate, the number of positive and negative words (sentiment score), and achieved 0.66 precision, 0.63 recall, and a 0.64 F-measure. Through manual error analysis, we noticed the importance of considering more comprehensive feature vector, including the number of positive and negative words to avoid some outrageous errors. Therefore, we enrich our feature vector by leveraging a LIWC program to glean linguistic styles signals, including auxiliary verbs, conjunctions, adverbs, functional words, and prepositions as well as number of positive and negative words. Surprisingly, these features only improved our F-measure by around 1%. Apart from that, we also used a distributed vector representation of training instances obtained from a pre-trained word2vec model on Twitter and Google News instead of using a discrete/traditional representation. Unfortunately, the performance decreased. Finally, we achieved the best accuracy by using deep learning based model (convolutional neural network). We discuss each of the key challenges and how we addressed them next.

## 1. Fast-paced change in dataset

The foremost challenging part is creating a robust classification system to cope with the dynamic nature of tweets related to an election. The election is very active (or dynamic) since everyday people talk about new aspects of an election and candidates in the context of unfolding events. Therefore, important features used to classify sentiment may soon become irrelevant and new, emerging features would be neglected if we did not update the training set regularly. Furthermore, in a political domain, unlike many other domains, people mostly express their sentiment toward the candidates implicitly and without using sentiment words extensively [6, 21]. This phenomenon makes the situation worse and more challenging. Another factor that may exacerbate the problem is differentiating the transient important features from lasting or recurring ones. Those features may disappear and then reappear in the future [14]. In the context of an election, for example, this scenario may happen because of the temporal changes in what each candidate's supporters talk about. Given this non-stationary characteristic of the election, we may encounter a concept drift/dataset shift problem. That is, learning when the test and training data have a different distribution. In fact, most of the machine learning approaches assume an identical distribution for the training and test set, although the test/target environment changes over time in many real-world problems. This phenomenon is an important factor for selecting our classification model. Among the classification models, SVM is one of the most robust models for dataset shift.

All of these aforementioned challenges make active learning necessary. There are two possible models for active learning which are useful to our problem as shown in the Figure 2. Both of these models are expensive as they involve humans in the loop for the labor-intensive and time-consuming task of annotation. Annotation is even more challenging here due to both the short length of tweets and the inherent vagueness in the political tweets that require awareness of political context on the part of the annotators. A question may arise concerning why we have not used any unsupervised approach like a lexicon-based approach when the annotation part is so challenging and our annotated dataset rapidly becomes obsolete and outdated. The answer is that in political tweets people often do not use many sentiment words; hence, the performance of a lexicon-based method would be low. To test, when we employed the MPQA subjectivity lexicon [20] to capture the subjectivity of each tweet, the accuracy maxed out at 0.49.

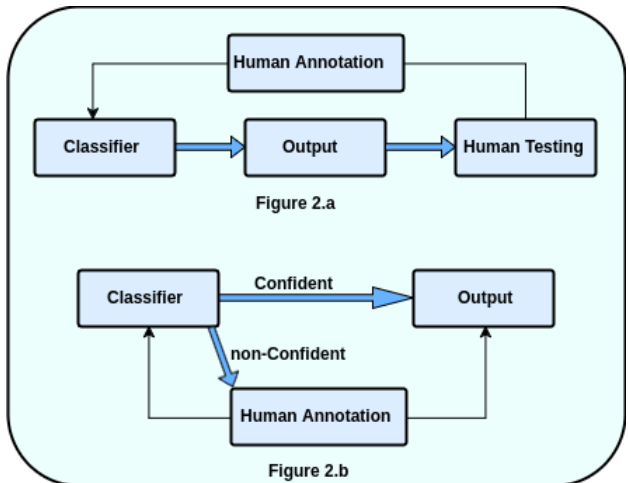

Figure 2. Active Learning Models.

Despite the costs, updating the training set regularly was the most effective measure for keeping the classifier reasonably good during the 2016 election. In fact, no matter how well our system may have been working, what worked well up until yesterday could become useless today after a new political event, set of propaganda, or scandal. For example, our first model trained during the primaries performed quite poorly during and after the conventions. Therefore, feeding the dataset with a new training set was the key task for keeping the system reliable. To do so, we were updating the training data in an opportunistic manner, usually every few days. It may also be worth trying to include more important/influential tweets in the training data. To achieve this goal we were collecting the data mostly at specific times, such as during the presidential debates.

**2. Candidate-dependence**

Most of sentiment analysis tools work in a target-independent manner. However, a target-independent sentiment analyzer is prone to yield poor results on our dataset because post-conventions a huge number of our tweets contain the names of both candidates. "I am getting so nervous because I want Trump to win so bad. Hillary scares me to death and with her America will be over" and "I don't really want Hillary to win but I want Trump to lose can we just do the election over" are examples of such tweets. Based on our observation, about 48% of our instances contained variants of both Clinton's and Trump's names. In such cases, the sentiment of those tweets may get misclassified for a given candidate because of the interference of the features related to another candidate. The state-of-the-art approaches for supervised target-dependent sentiment analysis can be grouped into two groups of syntax-based and context-based methods. The first group merely relies on POS tagging or syntax parsing for feature extraction (e.g., [9]), while the second group defines the left and right context for each target [18]. The latter outperforms the former in the classification of informal texts such as tweets [18]. To further enhance performance, sentiment lexicon expansion-related works such as [4] can be used to extract the sentiment-bearing candidate-specific expressions and those expressions can be added to the feature vector of a classifier. In our case, since we have trained one classifier per candidate, we can include the instances containing the name of more than one candidate in the training set of both classifiers. The key is to include features related to the target candidate in the corresponding classifier and exclude irrelevant ones both in the training and testing phase. To do that, we can use either dependency or proximity (similar to the two aforementioned works) to include the on-target features and ignore the off-target ones. Similarly, in the testing phase, depending on the classifier, we should include and exclude some of the features from our feature vector.

### 3. The importance of identifying the user's political preference

The ultimate goal of sentiment analysis over political tweet streams is predicting election results. Hence, obtaining some information about the political preference of the users can provide more fine-grained sources of information to a political pundit or analyzer for insight. Inspired by [4], we have developed a simple but effective algorithm to categorize users into five groups of far left-leaning, left-leaning, far right-leaning, right-leaning, and independent users. The idea behind our approach is the tendency of users to follow others who have a similar political orientation as themselves. The more right/left-leaning a user follows, the more likely that the particular user is right/left-leaning. Our approach involved collecting a set of Twitter users with known political orientation, including all senators, congresspersons, and political pundits. Then, we estimated the probability that a user is left-leaning(right-leaning) by calculating the ratio of left-leaning (right-leaning) followees of a user to his/her total number of followers. Finally, we decide the political preference of a user by comparing the above ratio with a threshold. Gaining this information about users helps to improve the social media-based prediction of the election.

### 4. Content-related challenges (hashtags)

Recently, there has been a surge of interest in distant supervision, which is training a classifier on a weekly labeled training set [7]. In this method, the training data gets automatically labeled based on a set of heuristics. In the context of sentiment analysis, using the emoticons :) and :( and other similar emoticons as a positive and a negative label respectively is one way of using distant supervision. Hashtags are also widely used for different machine learning tasks such as emotion identification [19]. People use a plethora of hashtags in their tweets about the election. Due to the dynamic nature of the election domain, the quality, quantity, and freshness of labeled data plays a vital role in creating a robust classifier. It is therefore desirable to use popular hashtags that each candidate's supporters use as a weak label in our dataset. However, our analysis for the 2016 election showed that hashtags were widely used for sarcasm, so using popular hashtags for automatic labeling leads to incorrectly labeling instances. For example, through the election only 43% of tweets containing #Imwithher were positive for Clinton, while it was used sarcastically in 27% of tweets. Consequently, our experiments show that using those hashtags as a feature for our classifier will decrease accuracy rather than increase it.

### 5. Content-related challenges (links)

All existing techniques for tweet classifiers rely merely on tweet contents and ignore the content of the documents they point to through a URL. However, around 36% of the 2016 election tweets contain a URL to an external link. In the 2012 election [4], we noticed d that 60% of tweets from very highly engaged users contain URLs. Those links are crucial as without them often the tweet is incomplete and inferring the sentiment is impossible or difficult even for a human annotator. Therefore, our hypothesis is that incorporating the content, keywords or title of the documents that a URL points to as a feature will cause a gain in our performance. To the best of our knowledge, there is no work on tweet classification that expands tweets based on their URLs. However, link expansion has successfully been applied to other problems such as topical anomaly detection [3] and distant supervision [12].

## 6. Content-related challenges (sarcasm)

To date, many sophisticated tools and approaches have been proposed to deal with sarcasm. More recently, [15] employ deep neural network (pre-trained convolutional neural network) for identifying sentiment, emotion, and personality features for sarcasm detection. Looking closer at these works, they mostly focus only on detecting the sarcasm in the text and not on how to cope with it in the sentiment analysis task. This raises the interesting question about how sarcasm may or may not affect the sentiment of the tweets and how to deal with sarcastic tweets in both the training and prediction phases. Rillof et al. [16] have proposed an algorithm to recognize the common form of sarcasm which flips the polarity in the sentence. These kinds of polarity-reverser sarcastic tweets often express the positive (negative) sentiment in the context of a negative (positive) activity or situation. However, Maynard et al. [13] show that determining the scope of sarcasm in tweets is still challenging. In fact, the polarity of sarcasm may apply to part of a tweet or its hashtags but not necessarily the whole. As a result, dealing with sarcasm in the task of sentiment analysis is an open research issue worth more work. Based on our observation, 7% of Trump's tweets and 6% of Clinton's tweets are sarcastic. Among these sarcastic tweets, 39% and 32% of them were classified incorrectly by our system. In terms of the training set, our hypothesis is that excluding the sarcastic instances from the training set will remove the noise and improve the quality of our training set.

## 7. Interpretation-related challenges (Sentiment Analysis versus Emotion Analysis)

Study of sentiment has evolved to the study of emotions, which has finer granularity. Positive, negative, and neutral sentiments can be expressed with different emotions such as joy and love for positive polarity; anxiety and sadness for negative; and apathy for neutral sentiment. Our emotion analysis on who tweeted #IVOTED in the 2016 US presidential election showed that Trump had many more tweets and individuals expressing joyful emotion compared to Clinton . Though the sentiment analysis favored Hillary in the early hours, emotion analysis was showing better support for Trump. We considered emotion as a better criterion for predicting people's action like voting and usually there are significant emotional differences in the tweets which belong to the same polarity. This was key to our successful prediction of the 2016 election.

## 8. Interpretation-related challenges (Vote counting versus engagement counting)

Most/all of the aforementioned challenges affect the quality of our sentiment analysis approach. It is also very important to correlate a user's online behavior and opinion with their actual vote. Chen et al. [4] show the more important role of highly engaged users in result prediction of the 2012 election. There are two plausible explanations for this. First, the more a user tweets, the more reliably we can predict his/her opinion. Second, highly active people are usually more influential and more likely to actually vote in the real world. That is why an election monitoring system should report both user-level normalized sentiment in addition to a tweet-level one. It is the end user analyzer's task to consider both of these factors in prediction.

## 9. The importance of location

An application that predicts the election result must consider each state's influence in the election by means of the number of electoral votes for that state. Many tools and approaches have been developed for both fine-grained [8] and coarse-grained [17] location identification in tweets for different purposes such as disaster management and election monitoring. In the latter case, the geographic location of a tweet or the user location in the profile can be used to estimate the user's approximate location. During the 2016 election, the spatial aspect of our Twitris system played a crucial role in assisting end users in predicting the election.

## 10. Trustworthiness-related challenges (Bots)

What happens when a large number of participants in a conversation are biased robots which artificially inflate social-media traffic by manipulating public opinion and spreading political misinformation? A social bot is a computer algorithm that automatically generates content over social media and is trying to emulate and possibly change public attitude. For the past few years, social bots have inhabited social media platforms. Similar to media reports (http://j.mp/BotE16, http://bit.ly/BotyE16). We also witness bot wars between the two sides. Research targeting pinpointing sources include use of supervised statistical models utilizing network features including retweets, mentions, and hashtag co-occurrence [11], user features (i.e., language, geographic locations, account creation time, number of followers, followees) [10], and timing features (i.e., content generation and consumption, by measuring tweet rate and inter-tweet time distribution) [22]. Our effort to identify the source that generates a tweet (checking whether it is originating from an API or not) used a hybrid and empirical approach gave fairly good results as discussed at http://bit.ly/E16KBot.

Acknowledgements: This work is supported in part by National Science Foundation (NSF) Award# IIP 1542911 "PFI:AIR-TT: Market-driven Innovations and Scaling up of Twitris - A System for Collective Social Intelligence." Cognovi Labs has licensed the Twitris technology, and Prof. Sheth is a cofounder of Cognovi Labs. We acknowledge significant participation of Cognovi Lab's James Mainord and Jeremy Brunn in US Election real-time analysis. Any opinions, findings, and conclusions or recommendations expressed in this material are those of the authors and do not necessarily reflect the views of the NSF.